\documentclass[10pt,twocolumn,letterpaper]{article}

\usepackage{cvpr}              
\definecolor{cvprblue}{rgb}{0.21,0.49,0.74}
\usepackage[pagebackref,breaklinks,colorlinks,allcolors=cvprblue]{hyperref}
\usepackage{multirow,multicol,graphicx}
\usepackage{listings}
\usepackage[table]{xcolor}
\def\confName{CVPR}
\def\confYear{2026}

\title{Temporal Inversion for Learning Interval Change in Chest X-Rays}

\author{Hanbin Ko\textsuperscript{1,2} \quad
Kyungmin Jeon\textsuperscript{1,2} \quad
Doowoong Choi\textsuperscript{3} \quad
Chang Min Park\textsuperscript{3,4,5}\\
\\
\parbox{\linewidth}{\small\centering
\textsuperscript{1}Interdisciplinary Program in Bioengineering, Seoul National University Graduate School\\
\textsuperscript{2}Integrated Major in Innovative Medical Science, Seoul National University Graduate School\\
\textsuperscript{3}Dept. of Radiology, Seoul National University Hospital\\
\textsuperscript{4}Seoul National University College of Medicine\\
\textsuperscript{5}Inst. of Medical and Biological Engineering, Seoul National University Medical Research Center\\
\scriptsize\tt lucasko1994@snu.ac.kr, kmjeon@snu.ac.kr,\\justicejustice314@snu.ac.kr, morphius@snu.ac.kr
}}

\begin{document}
\maketitle
\begin{abstract}
Recent advances in vision--language pretraining have enabled strong medical foundation models, yet most analyze radiographs in isolation, overlooking the key clinical task of comparing prior and current images to assess interval change. For chest radiographs (CXRs), capturing interval change is essential, as radiologists must evaluate not only the static appearance of findings but also how they evolve over time. We introduce \textbf{TILA} (Temporal Inversion-aware Learning and Alignment), a simple yet effective framework that uses \emph{temporal inversion}, reversing image pairs, as a supervisory signal to enhance the sensitivity of existing temporal vision--language models to directional change. TILA integrates inversion-aware objectives across pretraining, fine-tuning, and inference, complementing conventional appearance modeling with explicit learning of temporal order. We also propose a unified evaluation protocol to assess order sensitivity and consistency under temporal inversion, and introduce \textbf{MS-CXR-T\textsubscript{retrieval}}, a retrieval evaluation set constructed through a general protocol that can be applied to any temporal CXR dataset. Experiments on public datasets and real-world hospital cohorts demonstrate that TILA consistently improves progression classification and temporal embedding alignment when applied to multiple existing architectures.
\end{abstract}

\section{Introduction}
Vision language pretraining (VLP) models such as CLIP~\citep{clip} and SigLIP~\citep{siglip} have substantially advanced representation learning by aligning visual and textual information, achieving strong performance in retrieval and zero-shot classification. However, extending these methods to medical imaging is nontrivial, as clinical data are heterogeneous, reports are unstructured, and labeled image--text pairs are comparatively scarce~\citep{du2022survey,ko2025exploring}. These factors limit the direct transfer of general-domain VLPs to robust medical-image understanding tasks.

Recent medical VLPs have begun to bridge this gap by learning visual--textual alignment within the clinical domain, enabling disease classification and localization~\citep{convirt,huang2021gloria,biovil,wang2022medclip,wang2022multi}. Yet most existing models still analyze single CXRs in isolation, overlooking the core clinical task of comparing a patient’s current study with its most recent prior to assess interval change~\citep{zhu2023utilizing}. In routine interpretation, radiologists describe findings as \textit{improving}, \textit{stable}, \textit{worsening}, \textit{resolving}, or \textit{new}, emphasizing that temporal comparison is fundamental to diagnostic reasoning~\citep{hou2023recap}. Our focus aligns with standard radiology workflow, where each new examination is explicitly compared with the most recent prior.

\begin{figure*}[h]
    \centering
    \includegraphics[width=\linewidth]{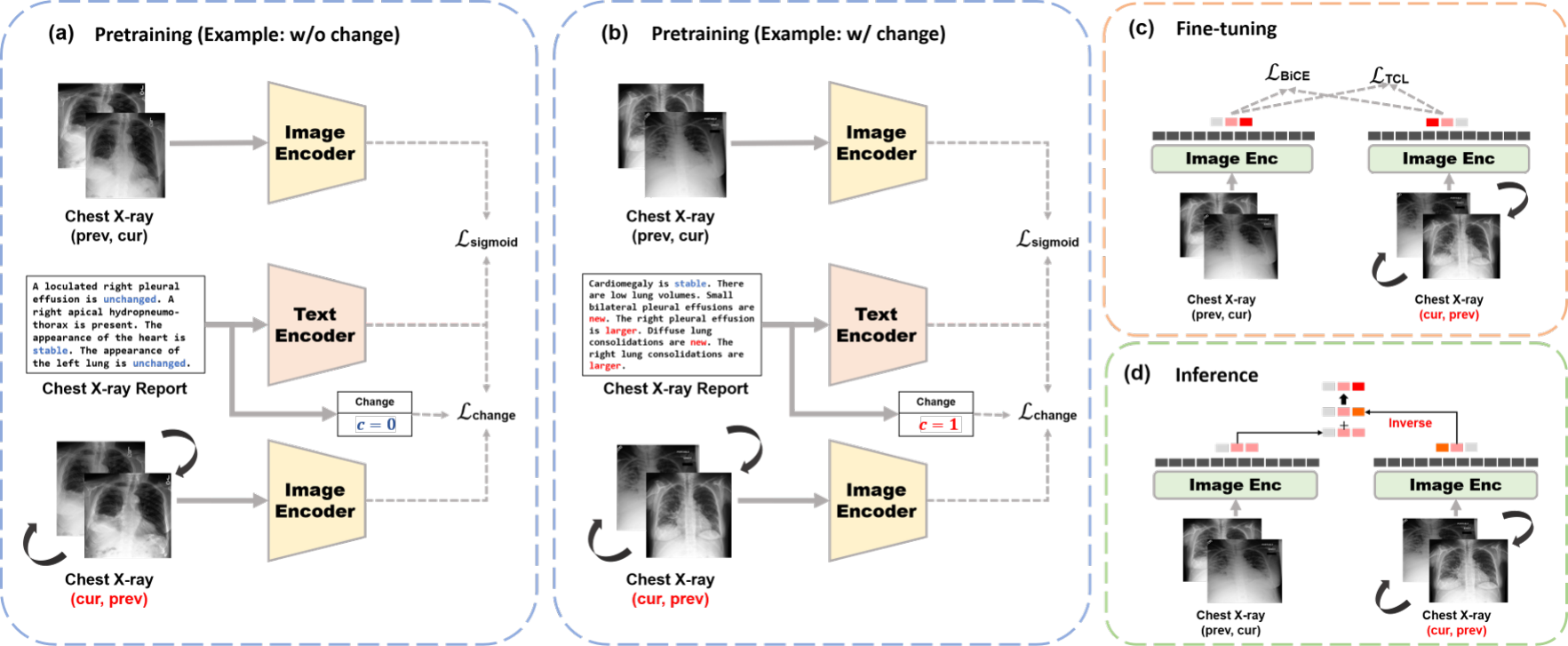}
    \caption{\textbf{TILA Framework: Temporal Inversion-aware Learning and Alignment.}
    The framework comprises three stages: pretraining, fine-tuning, and inference. 
    (a–b)~Pretraining: paired CXRs are encoded in both original and reversed orders; the Change-aware Sigmoid Loss aligns unchanged cases and separates changed ones. 
    (c)~Fine-tuning: the Bidirectional Cross-Entropy (BiCE) enforces label inversion, while the Temporal Consistency Loss (TCL) aligns probability distributions under reversal. 
    (d)~Inference: forward and reversed predictions are fused via inversion-aware scoring to enhance robustness and order consistency.}
    \label{fig:tila_framework}
\end{figure*}

To address this need, recent studies have introduced multi-image encoders for paired CXR analysis. BioViL-T~\citep{biovilt} incorporates prior images to represent temporal context, ALTA~\citep{alta} adapts masked vision models for efficient inter-study alignment, and TempA-VLP~\citep{tempavlp} employs contrastive objectives to learn dynamic changes across image pairs. Other approaches, including SDPL~\citep{sdpl}, CheXRelNet~\citep{karwande2022chexrelnet}, CheXRelFormer~\citep{chexrelformer}, and Med-ST~\citep{medst}, explore fine-grained or architectural mechanisms to capture anatomical and temporal patterns. Beyond image encoders, several methods leverage historical reports for temporal reasoning~\citep{huang2024hist}; while effective, these approaches depend on prior text availability and are thus complementary to image-based modeling. Notably, even the coarser task of binary interval-change screening, determining whether any change has occurred between studies, carries substantial clinical value, as it can prioritize changed cases for detailed review and reduce the reporting burden in high-volume longitudinal settings~\citep{yun2023deep}.

However, even with these temporal encoders, evaluation remains limited. Most studies rely on a single progression label, offering little insight into whether models truly capture the \textit{direction} of change. Directionality is central to diagnostic reasoning, yet progression labels are inherently noisy because temporal change is annotated without standardized criteria, making ``stable'' and borderline cases especially inconsistent. As a result, moderate classification scores do not necessarily indicate genuine temporal understanding and may reflect reliance on entity presence or chance agreement. To more directly probe directional reasoning, we examine model behavior under \emph{temporal inversion}: by reversing the image order and swapping corresponding labels, we can test whether predictions remain logically consistent across time directions. While such inversion is not always clinically symmetric, since not all recovery precisely mirrors worsening, many radiographic findings, particularly those involving changes in size, density, or extent, exhibit approximately reversible visual patterns (e.g., effusion volume, pneumothorax size, consolidation extent). This makes temporal inversion a useful analytical stress test for order sensitivity, providing stronger evidence of temporal reasoning beyond label priors.

To address these limitations, we propose \textbf{TILA}, a unified framework that leverages \textit{temporal inversion} as a supervisory signal for paired CXR analysis. TILA integrates inversion-aware objectives across pretraining, fine-tuning, and inference to learn directional and order-sensitive interval-change representations, and introduces \textbf{MS-CXR-T\textsubscript{retrieval}} for standardized evaluation of temporal reasoning.
Our work makes the following contributions:
\begin{itemize}
    \item We present \textbf{TILA}, a unified framework that incorporates temporal inversion throughout training and inference to learn order-aware interval-change representations in paired CXRs.
    \item We propose an \textbf{inversion-aware evaluation protocol} to quantify order sensitivity, inversion consistency, and bidirectional robustness, and introduce a reproducible pipeline for constructing progression-aware retrieval evaluation sets, instantiated as \textbf{MS-CXR-T\textsubscript{retrieval}}.
    \item We demonstrate through extensive experiments that TILA improves progression classification and temporal embedding alignment across multiple architectures and datasets, indicating generality and temporal robustness.
    \item We further show that TILA's temporal representations transfer to binary interval-change screening, a clinically important triage task.
\end{itemize}

\section{Method}
\label{sec:method}

We present \textbf{TILA}, a general framework that introduces order-awareness into temporal CXR analysis. Rather than modifying network architectures, TILA employs lightweight loss functions and an inversion-aware inference scheme that can be integrated into any paired-image vision--language backbone. We validate the framework on two representative temporal VLP models, BioViL-T~\citep{biovilt} and ALTA~\citep{alta}, demonstrating that the same principles consistently enhance directional reasoning across distinct architectures. The core idea is to leverage \textit{temporal inversion} as a supervisory signal throughout training and inference, encouraging models to more explicitly capture directional change and strengthen sensitivity to temporal order.

Formally, given a paired-image encoder $f_\theta$ and a text encoder $g_\phi$, TILA introduces inversion-aware objectives at three stages: \textbf{pretraining}, \textbf{fine-tuning}, and \textbf{inference}, as illustrated in \cref{fig:tila_framework}.

\subsection{Pretraining: Temporal Inversion with SigLIP}

\paragraph{Objective.}
The goal of pretraining is to learn temporally aware image--text representations by leveraging inversion as a contrastive signal. 
TILA introduces a \textbf{Change-aware Sigmoid loss}, built upon the SigLIP formulation~\citep{siglip}, which distinguishes whether a report describes interval change and whether new or resolved findings appear between exams. 
Thus, the alignment between an image pair and its associated report depends on the presence, absence, or direction of temporal change.

The key principle is that image--text alignment should behave differently depending on the underlying temporal relationship:
\begin{itemize}
    \item When the report indicates \textbf{no change}, both the original and inverted image pairs should align with the report, as the description remains valid regardless of temporal order.
    \item When the report indicates a \textbf{change}, the inverted pair should not align because the temporal direction is reversed and the described progression no longer holds true.
\end{itemize}
This design enables the model to capture the directionality of progression and distinguish genuine temporal changes from stability. Even in clinically irreversible cases (e.g., post-surgical or fibrotic findings), the inverted pair should not align with the report, since the reversal no longer represents a valid or clinically meaningful relationship between the images and the described change.

\paragraph{Image and Text Embedding.}
Let $\mathcal{B} = \{(x_i^{\text{prev}}, x_i^{\text{cur}}, r_i, c_i)\}_{i=1}^{|\mathcal{B}|}$ denote a mini-batch of paired CXRs, their corresponding reports, and binary change labels $c_i \!\in\! \{0,1\}$, where $c_i{=}1$ indicates \textbf{change} and $c_i{=}0$ indicates \textbf{no change}. The image encoder $f$ and text encoder $g$ produce:
\[
v_i = f(x_i^{\text{prev}}, x_i^{\text{cur}}), \quad 
v_i^{\text{swap}} = f(x_i^{\text{cur}}, x_i^{\text{prev}}), \quad 
t_i = g(r_i),
\]
where $v_i$ and $v_i^{\text{swap}}$ represent embeddings from the original and inverted image orders, and $t_i$ is the report embedding. The binary label $c_i$ is automatically assigned using an LLM that compares each follow-up report with its corresponding prior to identify interval changes, detecting both progression-related phrases (e.g., ``improved'', ``worsened'') and appearance or resolution of findings (e.g., ``new opacity'', ``no longer seen'').

\paragraph{Change-aware Sigmoid Loss.}
TILA builds upon the SigLIP pairwise sigmoid contrastive loss~\citep{siglip}, which aligns matched image--text pairs while repelling unmatched ones using learnable temperature and bias terms. We extend this with a \textbf{Change-aware Sigmoid Loss} applied to inverted image pairs $(x^{\text{cur}}, x^{\text{prev}})$, reweighting alignments according to the change label:
\begin{equation}
\label{eq:change_aware_loss}
\mathcal{L}_{\text{change}}
= -\frac{1}{|\mathcal{B}|}\!\sum_{i=1}^{|\mathcal{B}|}\!\sum_{j=1}^{|\mathcal{B}|}
\log \,\sigma\!\big(z^{\text{swap}}_{ij}(\tau^{\text{swap}}\, v_i^{\text{swap}\top} t_j - b^{\text{swap}})\big),
\end{equation}
with
\[
z^{\text{swap}}_{ij} =
\begin{cases}
+1, & (i{=}j)\ \&\ (c_i{=}0),\\[3pt]
-1, & (i{\neq}j)\ \text{or}\ (c_i{=}1),
\end{cases}
\]
and learnable parameters $\tau^{\text{swap}}$, $b^{\text{swap}}$. The key difference from standard SigLIP is that $z^{\text{swap}}_{ij}{=}{+}1$ \emph{only} when the example matches its own report ($i{=}j$) and the case is \textit{no change} ($c_i{=}0$); all other pairs---including matched pairs describing change---are treated as negatives.

\paragraph{Total Loss.}
The overall pretraining objective integrates the base and change-aware contrastive terms:
\begin{equation}
\label{eq:total_loss}
\mathcal{L}_{\text{total}}
= \mathcal{L}_{\text{siglip}} + W\,\mathcal{L}_{\text{change}},
\end{equation}
where $\mathcal{L}_{\text{siglip}}$ is the base pairwise sigmoid loss applied to original-order pairs~\citep{siglip} and $W$ controls the relative contribution of the change-aware component.
This joint objective yields direction-sensitive representations: stable cases align in both orders, whereas changed cases are discouraged from aligning under reversal.

\subsection{Fine-tuning: Enforcing Temporal Inversion Awareness}

\paragraph{Objective.}
Fine-tuning enforces temporal consistency during interval-change classification. Each paired CXR is labeled as $\texttt{improved}$, $\texttt{stable}$, or $\texttt{worsened}$. When the temporal order is reversed, the label should invert accordingly (e.g., $\texttt{improved} \leftrightarrow \texttt{worsened}$), while $\texttt{stable}$ remains unchanged. 

\paragraph{Bidirectional Cross-Entropy Loss (BiCE).}
Let $y \in \{\texttt{improved}, \texttt{stable}, \texttt{worsened}\}$ denote the ground-truth progression label for a paired input $(x^{\text{prev}}, x^{\text{cur}})$.  
We define the inversion mapping $\mathcal{I}(y)$ as
\[
\mathcal{I}(y) =
\begin{cases}
\texttt{worsened}, & \text{if } y=\texttt{improved},\\
\texttt{stable}, & \text{if } y=\texttt{stable},\\
\texttt{improved}, & \text{if } y=\texttt{worsened}.
\end{cases}
\]
The Bidirectional Cross-Entropy loss averages the classification losses from the forward and reversed orders:
\begin{equation}
\label{eq:bice_loss}
\begin{split}
\mathcal{L}_{\text{BiCE}}
= \tfrac{1}{2}\Big[
    \operatorname{CE}\!\big(f_\theta(x^{\text{prev}}, x^{\text{cur}}), y\big)
    \\
    +\operatorname{CE}\!\big(f_\theta(x^{\text{cur}}, x^{\text{prev}}), \mathcal{I}(y)\big)
\Big].
\end{split}
\end{equation}
This enforces label-level inversion consistency, treating progression as a directional continuum rather than a set of independent classes.

\paragraph{Temporal Consistency Loss (TCL).}
While BiCE enforces label-level symmetry, the \textbf{Temporal Consistency Loss} ensures that the probability distributions themselves transform consistently to inversion.  
Let $p_{\text{fwd}}$ and $p_{\text{bwd}}$ denote the predicted class probabilities for the forward and reversed orders, respectively.  
We define a transformation $\mathcal{S}$ that swaps probabilities between $\texttt{improved}$ and $\texttt{worsened}$ while keeping $\texttt{stable}$ fixed:

\begin{equation}
\label{eq:s_mapping}
\begin{aligned}
\mathcal{S}(p)[:,0] &= p[:,2], \\
\mathcal{S}(p)[:,1] &= p[:,1], \\
\mathcal{S}(p)[:,2] &= p[:,0].
\end{aligned}
\end{equation}
The TCL objective is then
\begin{equation}
\label{eq:tcl_loss}
\mathcal{L}_{\text{TCL}}
= \frac{1}{|\mathcal{B}|}\sum_{i=1}^{|\mathcal{B}|}
\left\| p_{\text{fwd}}^{(i)} - \mathcal{S}(p_{\text{bwd}}^{(i)}) \right\|^2.
\end{equation}
This encourages the model to assign mirrored probabilities under inversion, enforcing probability-level symmetry between forward and reversed predictions.
For instance, a case predicted as highly $\texttt{improved}$ in the forward order should correspondingly yield a high $\texttt{worsened}$ probability in the reversed order.

\paragraph{Total Fine-tuning Loss.}
The final fine-tuning objective combines classification and probabilistic consistency:
\begin{equation}
\label{eq:finetune_total}
\mathcal{L}_{\text{total}}
= \mathcal{L}_{\text{BiCE}} + \lambda\,\mathcal{L}_{\text{TCL}},
\end{equation}
where $\lambda$ balances accuracy and temporal coherence.  
Jointly optimizing BiCE and TCL enables the model to maintain order-aware and temporally coherent predictions across all interval-change categories.

\subsection{Inference: Inversion-aware Scoring}
\label{sec:inference}

\paragraph{Objective.}
During inference, TILA enhances robustness by aggregating predictions from both temporal directions.  
Given a paired input $(x^{\text{prev}}, x^{\text{cur}})$, the final prediction is obtained by averaging the forward prediction with the inversion-adjusted reversed prediction:
\begin{equation}
\label{eq:inversion_aware_score}
\operatorname{score}
= \tfrac{1}{2}\!\left[
p\!\big(f_\theta(x^{\text{prev}}, x^{\text{cur}})\big)
+ 
\mathcal{S}\!\big(p(f_\theta(x^{\text{cur}}, x^{\text{prev}}))\big)
\right],
\end{equation}
where $p(\cdot)$ denotes the predicted class probabilities, and $\mathcal{S}$ swaps the probabilities for \texttt{improved} and \texttt{worsened} while keeping \texttt{stable} unchanged.

This inversion-aware fusion reduces order bias and prediction variance, producing more consistent and temporally robust interval-change assessments.

\section{Experiments}
\label{sec:experiment}

We comprehensively evaluate the proposed \textbf{TILA} framework across multiple temporal vision--language tasks, including retrieval, zero-shot classification, supervised fine-tuning, and interval-change screening, benchmarking against state-of-the-art temporal VLPs and our retrained baselines. We also introduce \textbf{MS-CXR-T\textsubscript{retrieval}}, a evaluation set constructed through a reproducible pipeline, and evaluate generalizability on external datasets, including CheXpert~\citep{chexpertplus}, RexGradient~\citep{zhang2025rexgradient}, and a private cohort.

\subsection{Model Implementation}
TILA is implemented on two representative temporal vision--language backbones, \textbf{BioViL-T}~\citep{biovilt} and \textbf{ALTA}~\citep{alta}. 
To ensure fair comparison and improved batch efficiency, we retrain both base architectures using the SigLIP loss~\citep{siglip} instead of the original CLIP loss, following the same pretraining settings reported in their respective works. 
This re-trained variant (\textit{SigLIP}) serves as our baseline, upon which \textbf{TILA} introduces inversion-aware objectives without modifying the underlying architectures. 
The text encoder $g_\phi$ is initialized from pretrained CXR-BERT~\citep{biovilt}. 
Training proceeds in two stages, with inversion-aware objectives activated after an initial warm-up phase; introducing them from the start can cause the model to collapse toward predicting all cases as \texttt{stable}, since this label is trivially consistent under inversion:
\begin{itemize}
    \item \textbf{Pretraining:} Models are trained for 10 epochs using the standard SigLIP loss, followed by 20 epochs adding the \emph{Change-aware Sigmoid Loss} (30 epochs total). We set $W=1$ (equal weighting for the change-aware term), use a learning rate of $1\times10^{-4}$, batch size 144, and the AdamW optimizer.
    \item \textbf{Fine-tuning:} For progression classification, models are trained for 20 epochs with the BiCE loss and 30 additional epochs with the Temporal Consistency Loss (50 epochs total). 
    In full fine-tuning, all parameters are updated. 
    We set $\lambda=50$ to balance the scale between the Temporal Consistency and cross-entropy losses. AdamW is used for the optimizer.
\end{itemize}

Full hyperparameter and training details are provided in Appendix \cref{subsec:model}.

\begin{table*}[t]
\centering
\setlength{\tabcolsep}{8pt}
\caption{\textbf{Single vs.\ paired-image retrieval on MIMIC and CheXpert.}
Retrieval performance (Recall@k for I2T/T2I) and TEM score.
Single-image models process only the current image, whereas paired-image models encode (\textit{prev},\,\textit{cur}). 
For each backbone, \textbf{SigLIP} denotes retraining with the sigmoid contrastive loss (baseline), and \textbf{TILA} adds the Change-aware Sigmoid Loss. CheXpert results are reported as mean and 95\% CI across ten random subsamples of 3{,}000 paired CXRs.}
\newcolumntype{G}{>{\columncolor{black!5}}c}
\label{tab:retrieval}
\resizebox{\textwidth}{!}{
\begin{tabular}{lccccccG|ccccccG}
\toprule
\multirow{3}{*}{\textbf{Model}} 
    & \multicolumn{7}{c}{\textbf{MIMIC Retrieval}} 
    & \multicolumn{7}{c}{\textbf{CheXpert Retrieval}} \\
\cmidrule(lr){2-8} \cmidrule(lr){9-15}
 & \multicolumn{3}{c}{I2T} & \multicolumn{3}{c}{T2I} & \textbf{TEM} 
 & \multicolumn{3}{c}{I2T} & \multicolumn{3}{c}{T2I} & \textbf{TEM} \\
\cmidrule(lr){2-4} \cmidrule(lr){5-7} \cmidrule(lr){9-11} \cmidrule(lr){12-14}
 & @1 & @5 & @10 & @1 & @5 & @10 & 
 & @1 & @5 & @10 & @1 & @5 & @10 & \\
 \midrule
 \multicolumn{15}{l}{\textit{Single-image VLPs}} \\
Ko \& Park.\,\citep{bc2c} 
    & 6.6 & 23.2 & 33.8 & 8.4 & 24.5 & 34.7 & 7.4
    & \(6.2\pm0.1\) & \(16.1\pm0.2\) & \(24.7\pm0.1\) & \(6.6\pm0.2\) & \(15.9\pm0.2\) & \(23.0\pm0.2\) & \(5.7\pm0.4\) \\

BioViL-T (single image)~\citep{biovilt} 
    & 3.1 & 10.7 & 17.9 & 3.7 & 11.3 & 18.4 & 3.5
    & \(2.4\pm0.1\) & \(8.9\pm0.2\) & \(14.8\pm0.2\) & \(2.6\pm0.2\) & \(7.6\pm0.1\) & \(13.2\pm0.1\) & \(4.2\pm0.3\) \\

ALTA (single image)~\citep{alta} 
    & 6.5 & 18.2 & 27.2 & 11.0 & 28.7 & 41.3 & 9.0
    & \(6.3\pm0.1\) & \(17.9\pm0.2\) & \(26.1\pm0.2\) & \(8.2\pm0.2\) & \(22.7\pm0.1\) & \(32.1\pm0.1\) & \(10.2\pm0.3\) \\

BiomedCLIP~\citep{biomedclip} 
    & 0.4 & 2.0 & 3.1 & 0.6 & 2.3 & 3.2 & 1.6
    & \(0.2\pm0.1\) & \(1.7\pm0.2\) & \(2.8\pm0.3\) & \(0.6\pm0.2\) & \(2.2\pm0.2\) & \(3.3\pm0.3\) & \(1.9\pm0.3\) \\

\midrule
\multicolumn{15}{l}{\textit{Paired-image VLPs}} \\
BioViL-T~\citep{biovilt} 
    & 5.2 & 17.1 & 25.1 & 5.5 & 16.3 & 24.0 & 12.1 
    & \(3.1\pm0.1\) & \(9.8\pm0.1\) & \(15.2\pm0.1\) & \(2.8\pm0.1\) & \(9.2\pm0.2\) & \(14.2\pm0.1\) & \(13.7\pm0.2\) \\

BioViL-T\textsubscript{SigLIP} 
    & 11.4 & \textbf{35.1} & 46.1 & 11.8 & 34.2 & 46.9 & 15.7 
    & \(11.7\pm0.2\) & \(\textbf{31.0}\pm0.3\) & \(\textbf{40.6}\pm0.2\) & \(10.1\pm0.1\) & \(\textbf{26.4}\pm0.2\) & \(\textbf{35.8}\pm0.2\) & \(18.5\pm0.2\) \\

BioViL-T\textsubscript{TILA} 
    & 12.8 & 34.8 & 45.4 & 12.0 & 33.0 & 47.3 & \textbf{17.1} 
    & \(\textbf{12.5}\pm0.2\) & \(30.2\pm0.3\) & \(40.2\pm0.3\) & \(\textbf{10.7}\pm0.2\) & \(26.3\pm0.3\) & \(35.4\pm0.2\) & \(\textbf{20.8}\pm0.1\) \\

ALTA~\citep{alta} 
    & 9.1 & 24.2 & 34.9 & 11.8 & 32.1 & 43.8 & 11.6 
    & \(8.5\pm0.2\) & \(23.3\pm0.1\) & \(32.8\pm0.2\) & \(9.2\pm0.2\) & \(24.6\pm0.2\) & \(34.5\pm0.1\) & \(17.6\pm0.1\) \\

ALTA\textsubscript{SigLIP}  
    & 13.6 & 34.9 & \textbf{46.7} & 13.4 & \textbf{34.9} & \textbf{48.0} & 14.7 
    & \(8.4\pm0.3\) & \(23.4\pm0.3\) & \(33.4\pm0.2\) & \(8.2\pm0.3\) & \(23.8\pm0.1\) & \(34.9\pm0.1\) & \(17.8\pm0.2\) \\

ALTA\textsubscript{TILA}  
    & \textbf{13.8} & 34.4 & 45.8 & \textbf{13.8} & 34.1 & 47.2 & 16.0 
    & \(8.7\pm0.2\) & \(24.2\pm0.2\) & \(34.1\pm0.2\) & \(9.3\pm0.2\) & \(24.9\pm0.2\) & \(35.2\pm0.2\) & \(19.2\pm0.2\) \\

\bottomrule
\end{tabular}
}
\end{table*}
\subsection{Dataset Description}

\paragraph{Training Dataset.}
We train primarily on MIMIC-CXR~\citep{mimic3}, excluding images without prior counterparts and those overlapping with MS-CXR-T~\citep{biovilt}. Binary change/no-change labels for pretraining are generated using Gemini~2.0 Flash~\citep{gemini}. For classification, we use progression labels from Chest ImaGenome~\citep{chestimagenome} to perform 3-class temporal classification, following official splits for consistency across training, validation, and testing.

\paragraph{MS-CXR-T Dataset.}
MS-CXR-T contains paired CXRs from MIMIC annotated by radiologists as \texttt{improved}, \texttt{stable}, or \texttt{worsened}. It serves as the main benchmark for interval-change modeling.

\paragraph{MS-CXR-T\textsubscript{retrieval} Construction Protocol.}
We describe a general pipeline for constructing progression-aware retrieval evaluation sets from any temporal CXR dataset with annotated findings. Given a paired CXR case, the pipeline reformulates its report into a set of directional variants for a specific finding, while neutralizing unrelated findings to isolate the target progression direction.
For example, for a ``stable'' pneumothorax case with the original report \textit{``Right pneumothorax is unchanged. Consolidation is worsening.''}, the pipeline generates three directional variants:
\textit{``Right pneumothorax is improved. Consolidation is present.''},
\textit{``Right pneumothorax is unchanged. Consolidation is present.''}, and
\textit{``Right pneumothorax is worsened. Consolidation is present.''}
This process isolates the target finding and its progression direction while preserving the natural co-occurrence of other findings, enabling realistic yet controlled evaluation of temporal reasoning.
We instantiate this protocol on MS-CXR-T to produce \textbf{MS-CXR-T\textsubscript{retrieval}}.

\paragraph{External Validation Sets.}
We evaluate generalization on three external datasets: (i) paired CXRs from CheXpert~\citep{chexpertplus} for retrieval and temporal embedding alignment, (ii) a private tertiary hospital cohort annotated using corresponding radiology reports, covering four findings for progression classification and binary interval-change screening, and (iii) paired CXRs from RexGradient~\citep{zhang2025rexgradient} for binary interval-change screening. Detailed data curation and labeling protocols along with data ethics are provided in Appendix~\cref{subsec:data}, and the binary screening task setup is described in Appendix~\cref{subsec:interval_binary}.

\begin{table*}[t]
\centering
\newcolumntype{G}{>{\columncolor{black!5}}c}
\caption{\textbf{Interval-change classification results on MS-CXR-T and a private cohort.}
We report macro-accuracy (\%) for five findings—\textbf{CON}: consolidation, \textbf{PE}: pleural effusion, 
\textbf{PNE}: pneumonia, \textbf{PTX}: pneumothorax, and \textbf{EDE}: edema—together with the average across findings for each evaluation protocol 
(\textbf{Standard}, \textbf{Reversed}, \textbf{Combined}, \textbf{Consistency}; see Section~\ref{sec:evaluation}). 
For retrieval and zero-shot classification, \textbf{TILA} models use only the Change-aware Sigmoid pretraining loss, while in supervised settings they additionally apply inversion-aware fine-tuning (BiCE and TCL). 
\textbf{SigLIP} denotes our re-trained baselines using the sigmoid contrastive loss, and \textbf{TILA} refers to the corresponding models augmented with inversion-aware objectives.}
\label{tab:main2}
\renewcommand{\arraystretch}{1.2}
\resizebox{\textwidth}{!}{%
\begin{tabular}{lcccccG|cccccG|cccccG|cccccG}
\toprule
\textbf{Model} 
& \multicolumn{6}{c}{\textbf{Standard}} 
& \multicolumn{6}{c}{\textbf{Reversed}} 
& \multicolumn{6}{c}{\textbf{Combined}} 
& \multicolumn{6}{c}{\textbf{Consistency}} \\
\cmidrule(lr){2-7} 
\cmidrule(lr){8-13} 
\cmidrule(lr){14-19} 
\cmidrule(lr){20-25}
 & CON & PE & PNE & PTX & EDE & Avg
 & CON & PE & PNE & PTX & EDE & Avg
 & CON & PE & PNE & PTX & EDE & Avg
 & CON & PE & PNE & PTX & EDE & Avg \\
\midrule
\multicolumn{25}{l}{\textbf{MS-CXR-T\textsubscript{retrieval}}} \\
BioViL-T~\citep{biovilt} & 49.1 & 42.8 & 38.3 & 33.1 & 51.8 & 43.1 & 42.8 & 51.7 & 45.1 & 36.2 & 48.2 & 44.8 & 53.7 & 54.3 & 50.8 & 37.1 & 59.8 & 51.2 & 23.4 & 21.9 & 21.7 & 10.1 & 29.2 & 21.3 \\
BioViL-T\textsubscript{SigLIP} & 48.9 & 58.3 & 55.6 & 34.4 & 53.6 & 50.2 & 51.7 & 53.2 & 61.0 & \textbf{38.1} & 50.7 & 51.0 & 55.6 & 60.0 & 64.5 & 39.0 & 58.1 & 55.5 & 32.8 & 35.3 & 40.1 & 14.2 & 41.0 & 32.7  \\
BioViL-T\textsubscript{TILA} & \textbf{55.3} & \textbf{58.7} & \textbf{58.1} & \textbf{40.3} & \textbf{58.0} & \textbf{54.1} & \textbf{55.9} & \textbf{59.0} & \textbf{61.5} & 37.0 & \textbf{56.5} & \textbf{54.0} & \textbf{58.8} & \textbf{62.4} & \textbf{65.9} & \textbf{46.5} & \textbf{62.5} & \textbf{59.3} & \textbf{39.8} & \textbf{39.9} & \textbf{43.0} & \textbf{22.8} & \textbf{44.0} & \textbf{37.9}  \\
ALTA~\citep{alta} & 46.8 & 33.9 & 36.0 & 31.1 & 40.4 & 37.7 & 44.2 & 47.4 & 43.1 & 37.9 & 41.5 & 42.9 & 54.0 & 54.4 & 60.9 & 35.4 & 51.3 & 51.2 & 18.4 & 11.0 & 9.3 & 4.7 & 10.5 & 10.8  \\
ALTA\textsubscript{SigLIP} & 45.7 & 43.1 & 42.7 & 29.1 & 45.4 & 41.2 & 45.2 & 47.8 & 46.7 & 35.5 & 44.0 & 43.9 & 54.5 & 55.8 & 59.3 & 33.2 & 53.1 & 51.2 & 21.9 & 23.4 & 25.5 & 7.6 & 31.5 & 22.0  \\
ALTA\textsubscript{TILA} & 49.2 & 45.9 & 45.5 & 35.4 & 46.4 & 44.5 & 49.8 & 50.2 & 48.9 & \textbf{38.1} & 50.7 & 47.6 & 56.5 & 57.2 & 62.3 & 42.1 & 55.8 & 54.8 & 30.3 & 26.4 & 26.8 & 15.2 & 38.6 & 27.5  \\
\midrule
\multicolumn{25}{l}{\textbf{Zero-shot (MS-CXR-T)}} \\
BioViL-T~\citep{biovilt} & 34.9 & 37.4 & 34.5 & 37.0 & 42.8 & 37.4 & 37.3 & 38.6 & 41.0 & 33.1 & 42.0 & 38.4 & 48.6 & 53.4 & 53.8 & 34.8 & 54.1 & 49.0 & 10.5 & 10.7 & 10.9 & 9.9 & 16.8 & 11.8 \\
BioViL-T\textsubscript{SigLIP} & 42.7 & \textbf{50.1} & 42.7 & 29.1 & \textbf{55.4} & 44.0 & \textbf{47.2} & 43.8 & 49.7 & 35.5 & 51.0 & 45.5 & 46.1 & 55.8 & 51.3 & 33.2 & 58.8 & 49.1 & 21.9 & 23.4 & 27.5 & 7.6 & \textbf{39.5} & 24.0 \\
BioViL-T\textsubscript{TILA} & \textbf{46.5} & 46.3 & \textbf{51.9} & 38.8 & 52.3 & \textbf{47.2} & 46.2 & 46.3 & 49.7 & 38.6 & \textbf{52.4} & 46.7 & \textbf{56.7} & \textbf{58.0} & \textbf{59.1} & 42.2 & \textbf{60.0} & \textbf{55.2} & \textbf{27.4} & \textbf{24.6} & \textbf{30.8} & \textbf{33.2} & 36.1 & \textbf{30.5} \\
ALTA~\citep{alta} & 44.6 & 41.5 & 41.8 & 32.7 & 44.4 & 41.0 & 43.7 & 44.0 & 50.8 & 38.7 & 40.3 & 43.5 & 50.2 & 48.5 & 53.4 & 36.8 & 56.2 & 49.1 & 20.4 & 19.7 & 21.1 & 9.0 & 17.7 & 17.6 \\
ALTA\textsubscript{SigLIP} & 45.2 & 45.8 & 45.1 & 34.1 & 49.6 & 44.0 & 45.4 & 48.2 & 48.8 & 38.5 & 45.2 & 45.3 & 51.8 & 54.9 & 56.8 & 38.2 & 56.1 & 51.6 & 21.4 & 20.8 & 18.2 & 10.3 & 20.6 & 18.3 \\
ALTA\textsubscript{TILA} & 45.6 & 46.7 & 50.1 & \textbf{39.1} & 50.1 & 46.4 & 46.1 & \textbf{50.1} & \textbf{52.1} & \textbf{39.2} & 51.3 & \textbf{47.8} & 54.2 & 57.6 & 58.9 & \textbf{43.2} & 59.4 & 54.7 & 26.4 & 23.8 & 28.2 & 29.0 & 34.6 & 28.4 \\
\midrule
\multicolumn{25}{l}{\textbf{Supervised (MS-CXR-T)}} \\
CheXRelFormer~\citep{chexrelformer} & 51.1 & 56.0 & 37.1 & 34.1 & 59.0 & 47.5 & 42.5 & 50.7 & 36.0 & 34.6 & 58.8 & 44.6 & 40.8 & 54.2 & 44.7 & 33.3 & 56.4 & 45.9 & 28.0 & 49.2 & 10.7 & 29.5 & 52.3 & 34.0 \\
CNN-TF~\citep{biovilt} & 46.6 & 55.2 & 41.3 & 35.2 & 61.2 & 47.9 & 37.8 & 39.4 & 32.1 & 22.7 & 47.1 & 35.9 & 48.3 & 57.1 & 44.0 & 37.9 & 60.1 & 49.5 & 27.4 & 25.2 & 27.6 & 14.0 & 34.1 & 25.7 \\
CNN-TF\textsubscript{TILA} & 48.6 & 55.8 & 42.9 & 37.5 & 60.7 & 49.1 & 43.6 & 49.7 & 38.5 & 37.6 & 61.7 & 46.3 & 52.2 & 57.8 & 49.2 & 39.7 & 62.4 & 52.3 & 36.1 & 40.5 & 44.2 & 26.6 & 52.8 & 40.1 \\
MLRG~\citep{mlrg} & 61.3 & 65.7 & 61.7 & 44.8 & 67.5 & 60.2 & 51.3 & 55.2 & 50.2 & 37.1 & 63.1 & 51.4 & 59.1 & 64.7 & 64.6 & 45.5 & 67.5 & 60.3 & 38.2 & 49.3 & 44.7 & 25.2 & 58.1 & 43.1 \\
BioViL-T~\citep{biovilt} & 58.1 & 65.7 & 64.7 & 42.9 & 68.9 & 60.1 & 47.5 & 62.3 & 53.5 & 36.9 & 64.9 & 53.1 & 59.2 & 64.2 & 66.3 & 44.5 & 68.4 & 60.6 & 38.2 & 46.6 & 46.8 & 22.7 & 56.3 & 42.2 \\
BioViL-T\textsubscript{SigLIP} & 61.3 & 67.5 & 64.2 & 43.0 & 69.3 & 61.1 & 52.6 & 61.4 & 56.1 & 32.6 & 63.8 & 53.3 & 60.4 & 65.6 & 63.0 & 42.2 & 67.0 & 59.7 & 41.3 & 40.7 & 45.2 & 16.6 & 53.6 & 39.5 \\
BioViL-T\textsubscript{TILA} & 63.3 & 66.7 & \textbf{67.1} & \textbf{52.3} & \textbf{70.9} & \textbf{64.1} & \textbf{64.5} & \textbf{65.8} & \textbf{66.6} & \textbf{51.2} & \textbf{70.3} & \textbf{63.7} & \textbf{63.7} & \textbf{66.9} & \textbf{67.6} & \textbf{51.0} & \textbf{68.8} & \textbf{63.6} & \textbf{60.2} & \textbf{58.3} & 57.8 & \textbf{43.1} & \textbf{67.4} & \textbf{57.4} \\
ALTA~\citep{alta} & 62.2 & 66.2 & 62.5 & 45.5 & 69.2 & 61.2 & 53.4 & 57.1 & 49.8 & 37.3 & 65.3 & 52.6 & 60.2 & 65.6 & 65.8 & 46.2 & 66.8 & 61.0 & 39.2 & 49.6 & 45.4 & 24.3 & 57.2 & 43.2 \\
ALTA\textsubscript{SigLIP} & 63.2 & 65.4 & 63.4 & 46.0 & 70.1 & 61.7 & 52.4 & 55.9 & 52.1 & 38.0 & 66.3 & 53.0 & 61.7 & 64.5 & 66.2 & 44.9 & 68.5 & 61.2 & 35.7 & 48.4 & 47.1 & 23.0 & 60.1 & 42.9 \\
ALTA\textsubscript{TILA} & \textbf{64.1} & \textbf{68.3} & 66.2 & 49.3 & 70.1 & 63.6 & 60.5 & 62.3 & 58.3 & 45.8 & 66.7 & 58.8 & 62.3 & 66.1 & \textbf{67.6} & 48.4 & 68.4 & 62.6 & 55.4 & 57.9 & \textbf{58.1} & 38.7 & 62.9 & 54.6 \\
\midrule
\multicolumn{25}{l}{\textbf{Supervised (Private)}} \\
CNN-TF~\citep{biovilt} & 41.0 & 47.5 & 35.9 & 31.2 & - & 38.9 & 33.8 & 40.4 & 30.6 & 31.1 & - & 34.0 & 43.7 & 49.0 & 36.1 & 32.3 & - & 40.3 & 21.9 & 22.6 & 13.8 & 12.9 & - & 17.8 \\
MLRG~\citep{mlrg} & 56.2 & 59.3 & 58.8 & 52.1 & - & 56.6 & 44.4 & 49.2 & 49.4 & 56.2 & - & 49.8 & 58.2 & 57.1 & 52.1 & 55.2 & - & 55.7 & 25.3 & 29.2 & 29.8 & 24.2 & - & 27.2 \\
CheXRelFormer~\citep{chexrelformer} & 47.9 & 48.4 & 41.3 & 38.9 & - & 44.2 & 41.8 & 49.2 & 36.2 & 33.0 & - & 40.1 & 49.7 & 48.3 & 41.7 & 34.6 & - & 43.6 & 26.5 & 43.4 & 19.0 & 26.2 & - & 28.8 \\
BioViL-T~\citep{biovilt} & 55.0 & 57.1 & 60.3 & 49.7 & - & 55.6 & 44.8 & 48.1 & 50.2 & 53.4 & - & 49.2 & 57.6 & 56.7 & 54.5 & 53.7 & - & 55.7 & 29.8 & 34.6 & 33.3 & 25.1 & - & 30.7 \\
BioViL-T\textsubscript{SigLIP} & 53.9 & 59.3 & 51.3 & 56.6 & - & 55.3 & 45.6 & 53.1 & 48.8 & 48.1 & - & 48.9 & 56.0 & 59.3 & 54.9 & 52.0 & - & 55.6 & 30.4 & 35.6 & 34.7 & 23.7 & - & 31.1 \\
BioViL-T\textsubscript{TILA} & \textbf{67.9} & \textbf{70.2} & 63.0 & \textbf{63.3} & - & \textbf{66.1} & \textbf{61.5} & \textbf{65.7} & 55.2 & \textbf{62.3} & - & \textbf{61.2} & \textbf{64.7} & \textbf{66.8} & 58.3 & \textbf{62.2} & - & \textbf{63.0} & \textbf{58.4} & \textbf{55.2} & 53.4 & \textbf{51.4} & - & \textbf{54.6} \\
ALTA~\citep{alta} & 59.2 & 61.3 & 61.5 & 48.5 & - & 57.7 & 45.1 & 47.6 & 51.3 & 44.9 & - & 47.3 & 58.5 & 56.2 & 55.9 & 51.1 & - & 55.5 & 39.2 & 49.6 & 45.4 & 24.3 & - & 39.7 \\
ALTA\textsubscript{SigLIP} & 61.1 & 60.7 & 62.2 & 51.3 & - & 58.9 & 52.2 & 50.2 & 53.2 & 44.3 & - & 50.0 & 61.5 & 58.9 & 57.4 & 52.6 & - & 57.6 & 45.7 & 51.3 & 47.4 & 26.0 & - & 42.6 \\
ALTA\textsubscript{TILA} & 65.8 & 68.3 & \textbf{64.2} & 60.1 & - & 64.6 & 60.2 & 64.2 & \textbf{55.8} & 56.3 & - & 59.2 & 62.5 & 65.9 & \textbf{59.2} & 59.3 & - & 61.8 & 55.7 & 53.3 & \textbf{54.2} & 48.8 & - & 53.0 \\ 
\bottomrule
\end{tabular}
}
\end{table*}

\subsection{Evaluation Protocol}
\label{sec:evaluation}

We evaluate retrieval performance using Recall@$k$ for image-to-text (I2T) and text-to-image (T2I) tasks, and report the Temporal Embedding Matching (TEM) score~\citep{biovilt}, computed as the F1 overlap of temporal terms between the reference and retrieved reports.  

For progression classification (zero-shot and fully fine-tuned), we adopt four complementary evaluation settings:

\begin{itemize}
    \item \textbf{Standard:} Macro-accuracy on original image pairs using the ground-truth labels \texttt{improved}, \texttt{stable}, and \texttt{worsened}. This serves as the primary evaluation reflecting clinical interpretation.
    \item \textbf{Reversed:} Accuracy after reversing image order and corresponding labels, used to assess the model’s directional sensitivity.
    \item \textbf{Combined:} Average of forward and reversed predictions using inversion-aware scoring (\cref{sec:inference}), providing a robustness check through bidirectional reasoning.
    \item \textbf{Consistency:} A prediction is considered correct only if the model outputs the correct label in both directions, indicating temporal reliability and reducing chance agreement.
\end{itemize}

While the \textbf{Standard} setting represents the clinically relevant metric, the additional evaluations (\textbf{Reversed}, \textbf{Combined}, and \textbf{Consistency}) serve as analytical tools to quantify order sensitivity and stability.  
The clinical utility of these extended evaluations is further discussed in Appendix~\ref{subsec:clinical_utility}.

\section{Results}

\begin{figure*}[t]
    \centering
    \includegraphics[width=\linewidth]{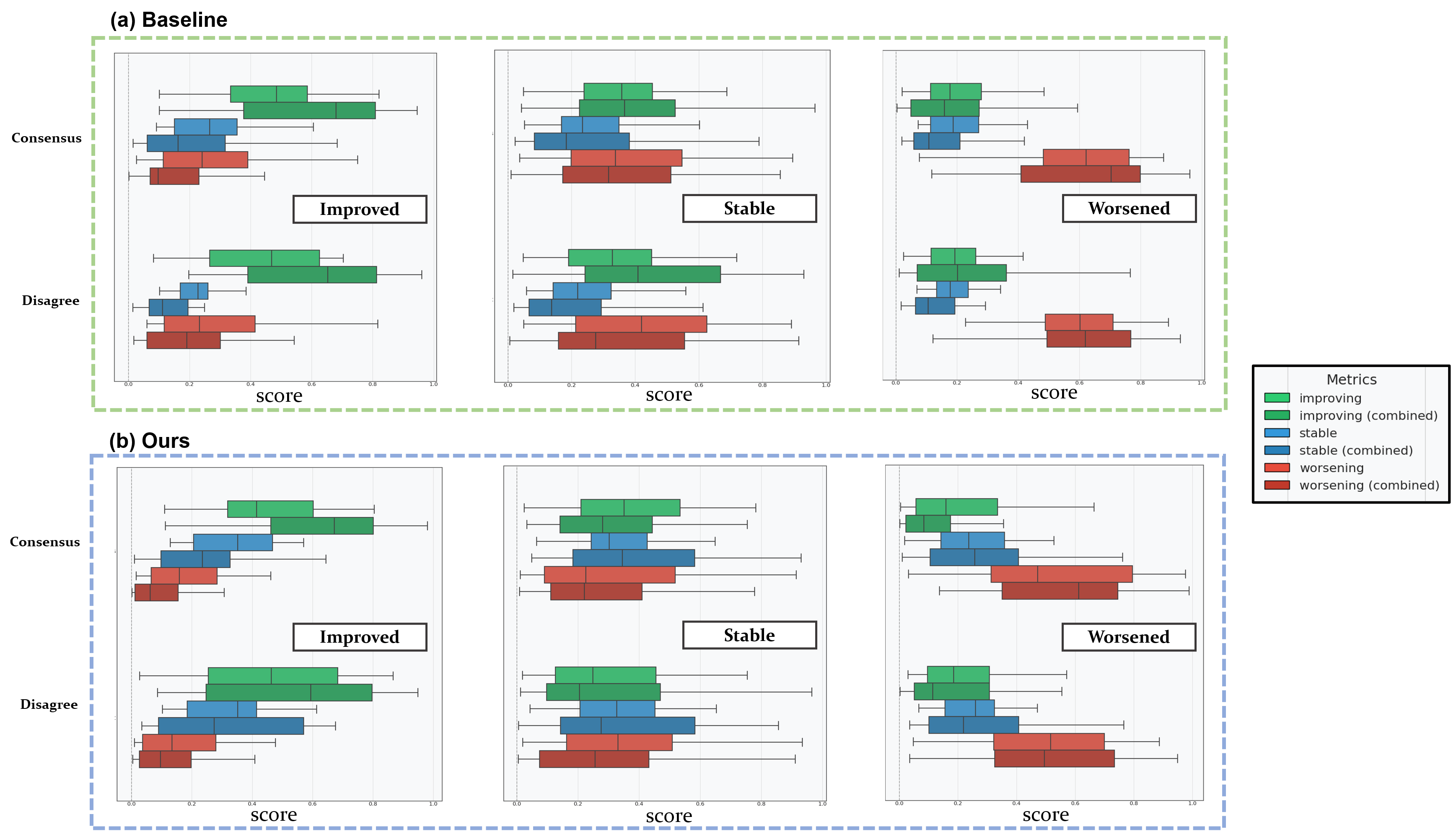}
    \caption{
    \textbf{Score Distribution Analysis for Pleural Effusion (MS-CXR-T).} 
    Distribution of prediction scores (scaled by 10) for each pleural effusion progression label, comparing baseline and TILA models in the zero-shot setting. Each boxplot separates cases by progression label (\texttt{improved}, \texttt{stable}, \texttt{worsened}) and label quality (\texttt{consensus} vs. \texttt{disagreement}), shown for both standard and inversion-aware (combined) scoring. Key trends illustrated:  
    i) Changes in scores for each label (\texttt{improved}, \texttt{stable}, \texttt{worsened}).  
    ii) Impact of applying inversion-aware scoring (combined).  
    iii) Differences in scores based on label quality (\texttt{consensus} vs. \texttt{disagreement}).  
    }
    \label{fig:tila_score}
\end{figure*}
\subsection{Experimental Results}

\paragraph{Retrieval Performance.}
As shown in \cref{tab:retrieval}, TILA yields comparable Recall to the base models while consistently improving TEM scores on both MIMIC and CheXpert, demonstrating that temporal inversion supervision enhances temporal alignment without compromising retrieval accuracy. Single-image baselines achieve similar Recall but substantially lower TEM (e.g., BioViL-T: 12.1 vs.\ 3.5 for the single-image), confirming that paired encoding provides critical temporal context that TILA further strengthens.

\paragraph{Interval-Change Retrieval Performance.}
On MS-CXR-T\textsubscript{retrieval} (\cref{tab:main2}), TILA improves the average Standard accuracy over SigLIP baselines for both backbones and yields consistent gains across the Reversed, Combined, and Consistency protocols, with particularly evident improvements for consolidation and pneumothorax. The inversion-aware combined scoring strategy provides modest additional boosts, likely by mitigating label ambiguities; this effect is examined further in the score analysis.

\paragraph{Zero-Shot and Supervised Performance.}
Across both zero-shot and fully supervised settings (\cref{tab:main2}), TILA consistently improves Standard Avg accuracy over the SigLIP baselines for both BioViL-T and ALTA. The gains are more pronounced in direction-sensitive protocols, with notably higher Consistency averages (e.g., zero-shot BioViL-T: 24.0\,\% $\rightarrow$ 30.5\,\%; supervised: 39.5\,\% $\rightarrow$ 57.4\,\%), indicating that inversion-aware supervision strengthens temporal coherence while maintaining or improving standard classification accuracy across architectures.

\paragraph{External Validation.}
On the external validation set from a private tertiary hospital (\cref{tab:main2}), the Avg columns show that models trained with TILA outperform their SigLIP baselines across Standard, Reversed, Combined, and Consistency protocols. These results suggest that inversion-aware supervision generalizes effectively beyond the training domain, contributing to more robust and temporally consistent performance in real-world clinical data.

\subsection{Ablation}

\begin{table}[h]
\centering
\caption{\textbf{Ablation of temporal inversion components.}
Each cell reports the mean macro-accuracy (\%) across the five findings
(CON, PE, PNE, PTX, EDE) for the indicated evaluation protocol.}
\label{tab:ablation}
\resizebox{\columnwidth}{!}{%
\begin{tabular}{lcccc}
\toprule
\textbf{Variant} & \textbf{Standard} & \textbf{Reversed} & \textbf{Combined} & \textbf{Consistency} \\
\midrule
BioViL-T (CLIP)               & 60.1 & 53.0 & 60.5 & 42.1 \\
+ SigLIP (Baseline)           & 61.1 & 53.3 & 59.6 & 39.5 \\
+ Change-aware SigLIP (Ours)  & 64.6 & 58.3 & 63.4 & 45.4 \\
\quad + BiCE                  & \textbf{65.2} & 62.0 & 63.4 & 53.1 \\
\quad + BiCE + TCL            & 64.1 & \textbf{63.7} & \textbf{63.6} & \textbf{57.4} \\
\bottomrule
\end{tabular}%
}
\end{table}

We perform ablation studies using BioViL-T as the backbone to isolate the contribution of each temporal inversion component. 
While the effects of inversion-aware pretraining on retrieval and zero-shot performance are shown in Tables~\ref{tab:retrieval} and~\ref{tab:main2}, here we focus on the supervised setting, where inversion-aware fine-tuning can be applied in addition to the Change-aware Sigmoid pretraining (\cref{tab:ablation}). 

Replacing the CLIP objective with the SigLIP loss produces modest shifts in the averaged metrics: Standard accuracy increases slightly (60.1\,\% $\rightarrow$ 61.1\,\%), while Combined and Consistency averages change minimally. 
Adding the Change-aware Sigmoid pretraining yields a larger improvement across all four protocols (Standard: 64.6\,\%; Reversed: 58.3\,\%; Combined: 63.4\,\%; Consistency: 45.4\,\%), indicating that inversion at pretraining provides a stronger initialization for downstream temporal classification.

Building on this initialization, incorporating BiCE during fine-tuning primarily boosts direction-sensitive metrics, raising the Reversed and Consistency averages while maintaining the Combined average and slightly improving the Standard average. 
Adding TCL on top trades a small reduction in Standard for further gains in Reversed, Combined, and especially Consistency, reinforcing temporal symmetry at the probability level. 

Additional hyperparameter analysis is provided in Appendix~\ref{sec:abl_hyperparameters}.

\subsection{Score Analysis}
To examine how TILA affects model behavior beyond classification accuracy, we analyze effusion predictions from MS-CXR-T with respect to label quality. Labels are categorized as \texttt{multiple experts}, \texttt{one-expert}, or \texttt{disagreement}. A \texttt{multiple experts} label indicates consensus among radiologists, whereas \texttt{disagreement} reflects variability in which one reader may assign \texttt{worsened} or \texttt{improved} while another assigns \texttt{stable}. These categories help contextualize expected uncertainty in the ground truth.

As shown in \cref{fig:tila_score}, inversion-aware scoring (\cref{sec:inference}) consistently shifts predictions toward the correct label, improving separation between classes and pushing incorrect scores further away. This effect is most pronounced in consensus cases, suggesting that integrating both temporal directions reduces ambiguity and sharpens calibration. For cases with \texttt{disagreement}, TILA produces smaller shifts, reflecting more conservative behavior when the underlying labels are uncertain—mirroring how radiologists temper confidence in ambiguous settings.

A notable improvement appears in distinguishing \texttt{stable} from \texttt{improved} and \texttt{worsened}. The baseline model frequently assigns low confidence to \texttt{stable}, causing overlap between progression categories. TILA widens this margin through inversion-aware scoring, yielding more reliable and well-separated stability scores. Clinically, stronger discrimination of stable cases is important for reducing unnecessary interventions and supporting more dependable longitudinal assessment.

\begin{table}[h]
\centering
\caption{Binary interval-change screening on two external test sets. AUC is reported for fine-tuning and linear probing modes.}
\label{tab:binary_screening}
\small
\begin{tabular}{llcc}
\toprule
\multirow{2}{*}{Model} & \multirow{2}{*}{Mode} & \multicolumn{2}{c}{Test Set AUC} \\
\cmidrule(lr){3-4}
 &  & Private & RexGradient \\
\midrule
BioViL-T                      & Fine-tune     & 0.734          & 0.565          \\
BioViL-T                      & Linear probe  & 0.693          & 0.545          \\
\addlinespace
BioViL-T\textsubscript{TILA} & Fine-tune     & \textbf{0.765} & \textbf{0.702} \\
BioViL-T\textsubscript{TILA} & Linear probe  & 0.756          & 0.681          \\
\bottomrule
\end{tabular}
\end{table}

\subsection{Binary Interval-Change Screening}
We evaluate whether TILA's representations generalize to binary change detection, determining whether any interval change occurred (\cref{tab:binary_screening}). BioViL-T\textsubscript{TILA} improves AUC over the baseline on both test sets and in both evaluation modes. In fine-tuning mode, TILA increases AUC from 0.734 to 0.765 on the Private test set and from 0.565 to 0.702 on RexGradient, demonstrating consistently better performance than the baseline across external datasets.

\section{Discussion and Conclusion}
\label{sec:discussion_conclusion}

Modeling interval change in paired CXRs remains a challenging problem. 
We introduced \textbf{TILA}, a framework that incorporates temporal inversion into pretraining, fine-tuning, and inference to enhance direction-sensitive reasoning. 
Across retrieval, zero-shot, and supervised evaluations, TILA improves temporal alignment metrics such as TEM and order-consistency metrics, while also yielding higher \textbf{Standard} accuracy for most findings and settings. 
The inversion-aware scoring strategy further stabilizes predictions, and TILA’s modular design enables seamless integration with existing vision--language pipelines.

\paragraph{Limitations.}
Inter-reader variability remains a fundamental challenge in interval-change labeling, particularly for subtle or borderline changes~\citep{schilling2022automated,nichyporuk2022rethinking,kang2022variability}. 
A further difficulty is the lack of unified criteria for defining progression across different findings, as visual manifestations of change vary widely and are not consistently standardized in clinical practice. 
Developing consensus definitions, along with curated multi-expert labels, would provide more reliable ground truth and strengthen the robustness of future temporal modeling approaches.


\paragraph{Asymmetry of recovery.}
Temporal inversion is not meant to represent a clinically symmetric reversal of disease, and its validity differs across TILA's stages. During \emph{pretraining}, swapped pairs with change are used only as negative supervision, making the objective valid regardless of clinical reversibility. For \emph{fine-tuning and inference}, label inversion assumes improved $\leftrightarrow$ worsened symmetry, which is not universally met; cautious interpretation is warranted for findings with inherently asymmetric trajectories. Nonetheless, inversion-aware supervision improves Standard accuracy, the clinically relevant metric, for most findings, indicating that it reinforces rather than disrupts forward-order predictions.

\paragraph{Conclusion.}
TILA provides a simple, architecture-agnostic approach for enhancing temporal sensitivity in paired CXR interpretation. 
By leveraging temporal inversion across training and inference, the framework improves directional alignment and order stability while maintaining or improving Standard accuracy in most settings. 
TILA offers a principled building block for future temporal vision--language models and may support more reliable interval-change assessment when combined with standardized labels and richer temporal context.

\section*{Acknowledgements}
We thank H.Y.\ Cho and K.J.\ Cho for their valuable advice on this work. This research was supported by a grant of the Korea Health Technology R\&D Project through the Korea Health Industry Development Institute (KHIDI), funded by the Ministry of Health \& Welfare, Republic of Korea (grant number: RS-2025-02213531, NTIS number: 2460004432), and by the National Research Foundation of Korea (NRF) grants funded by the Ministry of Science and ICT (MSIT) (Grant No.\ RS-2024-00354666).

{
    \small
    \bibliographystyle{ieeenat_fullname}
    \bibliography{main}
}

\clearpage
\setcounter{page}{1}
\maketitlesupplementary


\section{Model}
\label{subsec:model}
This section provides additional details on implementation, augmentations, hyperparameters, and model selection.

\subsection{Implementation Details}
For pretraining, we use the AdamW optimizer with a cosine learning-rate schedule and a 100-step warm-up. 
The learning rate is set to $1\times10^{-4}$ with a batch size of 144, and all parameters are trained in \texttt{bfloat16}. 
Pretraining is conducted on three NVIDIA A6000 GPUs for 30 epochs (approximately 54 GPU-hours).

Fine-tuning also uses AdamW with cosine scheduling and a 5\% warm-up. 
Models are trained end-to-end for 50 epochs using a learning rate of $1\times10^{-5}$ and a batch size of 128. 
These experiments are performed on a single NVIDIA A6000 GPU. 
Projection layers for both image and text encoders have dimension 128, and the CXR-BERT text encoder is configured with a maximum token length of 256.

\subsection{Augmentation}
Image augmentations follow the protocols described in BioViL-T~\cite{biovilt} and ALTA~\cite{alta} and are applied consistently across both pretraining and fine-tuning.

\subsection{Hyperparameters}
Logit scales $\tau$ and $\tau^{\text{swap}}$ are initialized to $\log 10$, with bias $-10$ (SigLIP’s logit-shift parameter). 
We set $W=1$ for the Change-aware Sigmoid loss during pretraining and choose $\lambda=50$ for fine-tuning to balance the magnitude of the Temporal Consistency Loss relative to the cross-entropy term. 
The Change-aware Sigmoid loss is activated after 10 pretraining epochs, and TCL is applied after 20 fine-tuning epochs to prevent premature convergence toward predicting \texttt{stable} for most pairs.

\subsection{Compared Models}
We compare only to models with publicly available code to ensure consistent training and reproducible evaluation. 
The CNN+Transformer baseline follows the BioViL-T~\cite{biovilt} implementation with ImageNet initialization. 
Methods without publicly available code or reproducible training pipelines were excluded.

Temporal CXR baselines remain limited relative to single-image CXR methods. 
Few existing approaches explicitly model interval change, and the field lacks standardized benchmarks, unified data splits, and publicly released training workflows. 
To avoid data leakage and ensure fair comparison, we include only models we can reproduce end-to-end and evaluate under our inversion-aware protocol.
Where available, we also include models with released pretrained weights, even if full training code is not provided. 
Recent studies similarly compare against a narrow set of baselines and often rely on reported numbers rather than independent reproduction. 
Given our focus on temporal robustness—not only raw accuracy—and the need to strictly avoid any MS-CXR-T overlap, controlled and fully reproducible evaluation was essential.

\section{Data}
\label{subsec:data}

This section describes dataset splits, label generation protocols, and ethical considerations, including specific details on the \textbf{MS-CXR-T\textsubscript{retrieval}} benchmark.

\subsection{Dataset Splits}
We first exclude chest X-ray images without available prior images. The sample counts for each split are presented in \cref{tab:split}. For CheXpert, since the official validation and test splits lack corresponding reports or prior images, we use the CheXpert training split as an external pool for retrieval evaluation. To minimize sampling bias, we repeatedly sample 3,000 image pairs from this pool 10 times and report the mean and 95\% confidence interval across these subsamples. For our private dataset, we collected CXR pairs from 2010–2020, filtering for reports and images containing temporal keywords (e.g., “improved”, “worsened”, “stable”). Labeling was performed by a researcher with four years of experience in chest X-ray interpretation, and these labels were used as the reference for external validation during fine-tuning.

\begin{table}[h!]
    \centering
    \begin{tabular}{|c|ccc|}
        \hline
        & Train & Validation & Test \\
        \hline
        MIMIC & 183,302 & 1,330 & 2,871 \\
        \hline
        CheXpert & - & - & 123,374 \\
        \hline
        Private & - & - & 2,233 \\
        \hline
    \end{tabular}
    \caption{Data Distribution for MIMIC, CheXpert, and Private}
    \label{tab:split}
\end{table}

\subsection{Dataset Approvals and Ethics}
All LLM-assisted label extraction for MIMIC and the construction of MS-CXR-T\textsubscript{retrieval} were conducted in accordance with PhysioNet guidelines for responsible LLM usage (https://physionet.org/news/post/gpt-responsible-use). 
All preprocessing for MS-CXR-T\textsubscript{retrieval} was performed \emph{prior to September~2025}, during a period in which de-identified report processing on Google Cloud Vertex~AI was permitted under the platform’s data-handling policies.

The private clinical dataset was collected under institutional IRB approval, and all researchers accessing the data were formally registered, credentialed, and authorized for handling de-identified records.

\subsection{Change Label Generation}
\label{subsec:changelabel}
Change/no-change labels for pretraining were generated using Gemini~2.0~Flash. 
The model compares each follow-up report with its corresponding prior (when available) and identifies interval changes, including \textit{improved/stable/worsened} and \textit{new/resolved} findings. 
For compatibility with the Change-aware Sigmoid Loss, we invert the binary output (1$\rightarrow$0, 0$\rightarrow$1), and exclude uncertain cases (``-1'') from training.

\paragraph{Sampled-scale validation of LLM choices.}
To verify that Gemini is not strictly required and to assess the reliability of alternative LLMs, we conducted a small-scale validation on 200 randomly sampled report pairs. 
For each pair, we validated the LLM-derived change/no-change label against a manually reviewed label provided by a trained annotator.

Agreement rates (with 95\% Wilson confidence intervals) were:

\begin{itemize}
    \item \textbf{Gemini~2.0~Flash:} 192/200 (96.0\%; 92.3–97.9)
    \item \textbf{Qwen3-14B:} 179/200 (89.5\%; 84.5–93.1)
\end{itemize}

These results show that smaller LLMs can also produce reasonable labels, although Gemini exhibited the highest agreement in this pilot study.
The prompt used for label generation is provided below:

\begin{scriptsize}
\begin{quote}
You are given two chest X-ray (CXR) reports: a **previous** CXR report and a **follow-up** CXR report. Your task is to analyze both reports and determine if there are any changes in the follow-up CXR compared to the previous one.

\#\#\# Instructions:
1. Compare the findings in both reports.
2. If there are **any new, worsening, or improving conditions**, return `1`.
3. If the reports state **"no interval change"** or findings are **stable**, return `0`.
4. If unsure, return '-1'.
5. Ensure the output is strictly `0`, `1`, '-1' without additional text.

\#\#\# Input Format:
- **Previous:** [Insert previous report text]
- **Follow-up:** [Insert follow-up report text]

\#\#\# Output Format:
Return only:
- `0` if no changes are detected.
- `1` if any changes (new, improved, or worsened findings) are detected.
- `-1` if uncertain, or any of the previous report or followup report is not a chest X-ray report.

\#\#\# Few-shot Examples:

\#\#\#\# **Example 1: No Changes (Output: 0)**
- **Previous:**  
  "Left pleural effusion is noted. The cardiac silhouette is normal. No acute abnormalities."
- **Follow-up:**  
  "No interval change."
- **Output:** `0`

\#\#\#\# **Example 2: New Finding (Output: 1)**
- **Previous:**  
  "The lungs are clear. No pleural effusion or pneumothorax. No focal consolidation. The cardiac silhouette is normal. No acute abnormalities."
- **Follow-up:**  
  "A new left lower lobe consolidation is noted, concerning for pneumonia. No pleural effusion or pneumothorax. The cardiac silhouette is normal."
- **Output:** `1`

\#\#\#\# **Example 3: Improvement in Findings (Output: 1)**
- **Previous:**  
  "Patchy bilateral infiltrates consistent with pneumonia. No pleural effusion. The heart size is within normal limits."
- **Follow-up:**  
  "Bilateral infiltrates have significantly improved. No pleural effusion. The heart size remains within normal limits.
- **Output:** `1`

\#\#\#\# **Example 4: Stable Findings (Output: 0)**
- **Previous:**  
  "Mild left basilar atelectasis. No pneumothorax or pleural effusion. No acute cardiopulmonary abnormalities."
- **Follow-up:**  
  "Mild left basilar atelectasis remains unchanged. No pneumothorax or pleural effusion."
- **Output:** `0`
\end{quote}
\end{scriptsize}
\begin{figure*}[h]
    \centering
    \includegraphics[width=\linewidth]{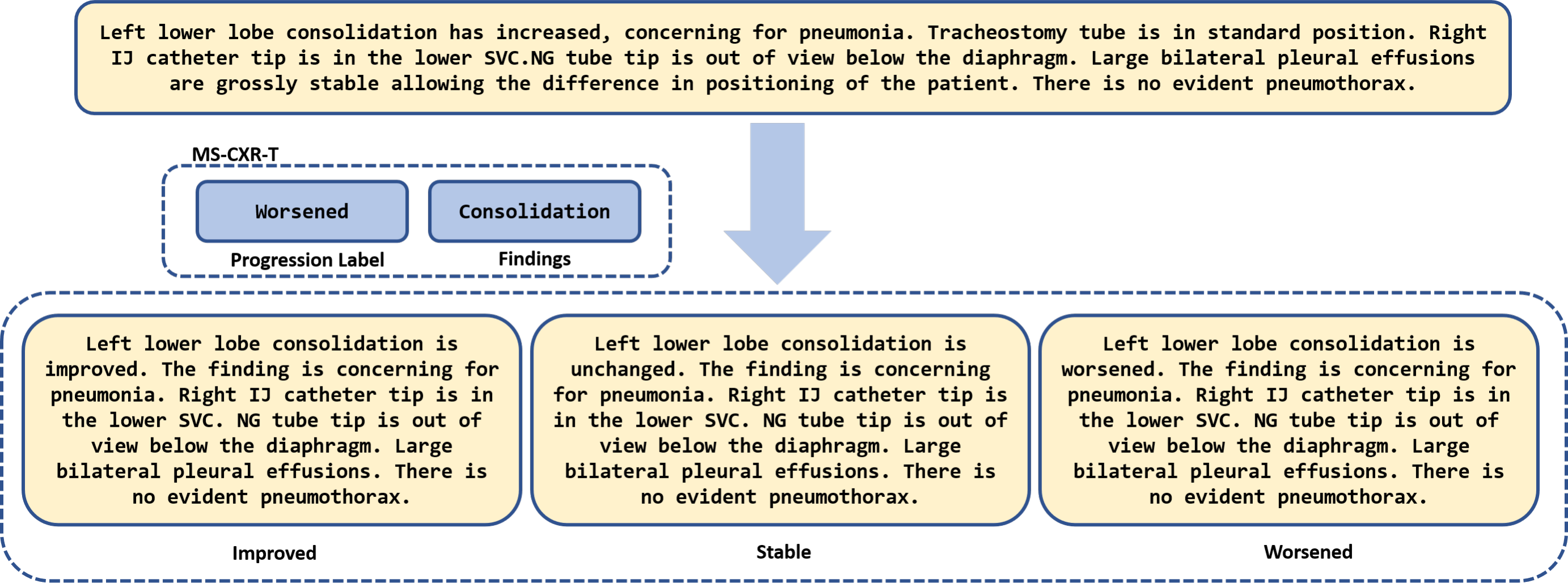}
    \caption{
    Example workflow for constructing MS-CXR-T\textsubscript{retrieval} from the original benchmark.}
    \label{fig:mscxr-T}
\end{figure*}

\subsection{MS-CXR-T\textsubscript{retrieval}}
\label{subsec:mscxrtretrieval}

We describe here the construction process of \textbf{MS-CXR-T\textsubscript{retrieval}}, our benchmark for evaluating temporal reasoning in CXR report retrieval. The creation pipeline consists of three main stages:

\textbf{Stage 1: Splitting Reports.}
We first extract the findings section and the corresponding report for each image. Reports are split so that each sentence describes a single radiological finding. This reduces noise in later stages when manipulating progression labels.
The prompt used for this sentence-level splitting is as follows:

\begin{scriptsize}
\begin{quote}
Analyze the given radiology report text and split it into sentences, each describing a single radiological finding or view position information. Include both positive and negative findings, as well as view position details, but exclude other non-finding information. Follow these guidelines:

1. Each sentence should contain only one of the following: (exclude the sentence with view position information)
   a) A clear radiological finding
   b) The absence of a specific condition (negative finding)

2. Treat each negative finding (absence of a condition) as a separate observation and split it into its own sentence.

3. Keep sentences that describe view position information, but separate them from findings if they appear in the same original sentence.

4. Exclude sentences that do not describe actual radiological findings or view positions, such as:
   - Procedural details
   - General comments about image quality
   - Patient positioning information (unless it's specifically about the view position)

5. Maintain the meaning and context of the original findings and view positions while splitting.

6. Minor sentence structure changes or addition of necessary words are allowed to ensure clarity.

7. Remove any redundant information and express each finding or view position concisely.

8. Each split sentence should be understandable independently.

9. Avoid using lists or enumerations within a single sentence; instead, create separate sentences for each item.

Example of splitting, including view position, and excluding non-findings:
Original Report: '1. A single frontal view of the chest is provided. 2. No consolidation, pleural effusion, or pneumothorax is observed in both lungs. 3. The heart size is normal.'
Output: 'No consolidation is observed in both lungs. No pleural effusion is observed in both lungs. No pneumothorax is observed in both lungs. The heart size is normal.'

Process the input text according to these guidelines and return the relevant radiological findings and view position information and do not attach any additional text except the split sentences.

\end{quote}
\end{scriptsize}

\textbf{Stage 2: Prior Reference Omission.}
We remove all references to prior studies from each finding. This neutralizes progression status for unrelated findings and allows our benchmark to focus evaluation on the target finding(s) defined in MS-CXR-T. This omission step also enables evaluation in reversed or swapped settings.
The prompt used for removing prior references is:
\begin{scriptsize}
\begin{quote}
You are an expert chest X-ray (CXR) radiologist familiar with radiologic reports. Your task is to rewrite the given radiology reports by removing all references to prior reports or comparisons, while preserving the original structure as much as possible.
Input: A radiology report for a chest X-ray (CXR).
Output: A revised CXR report focusing solely on current medical findings, excluding references to prior reports, comparisons, and irrelevant details.
Guidelines:
Remove Comparisons: Eliminate any terms or phrases that suggest a comparison, such as "compared to," "in comparison with," "change", "cleared", "constant", "decrease", "elevate", "expand", "improve", "decrease", "increase", "persistent", "reduce", "remove", "resolve", "stable", "worse", "new", etc.
Focus on Current Findings: Ensure the report only describes the current state of the patient's lungs and related structures.
Preserve Medical Context: Maintain the original medical terminology and descriptions of abnormalities.
Retain Negations: Keep any negative statements about the absence of abnormalities.

Example:
Original Report: The left apex has not been included on this radiograph. The ET tube terminates 3.9 cm above the carina. The NG tube terminates in the stomach. Surgical clips and a faint metallic coil project over the chest. A left PICC terminates in the mid SVC. EKG leads overlie the chest wall. The lung volumes are low. There are persistent bilateral mid and lower zone hazy opacities. There are persistent bilateral hilar and perihilar linear opacities. No significant interval change is observed in the lung opacities. Bilateral pleural effusions are present. The right pleural effusion is greater than the left. No pneumothorax is observed on the right. No cardiomegaly is present. No interval change is observed in the mediastinal silhouette. No significant interval change is observed in the bony thorax.  
Output: The left apex has not been included on this radiograph. The ET tube terminates 3.9 cm above the carina. The NG tube terminates in the stomach. Surgical clips and a faint metallic coil project over the chest. A left PICC terminates in the mid SVC. EKG leads overlie the chest wall. The lung volumes are low. There are persistent bilateral mid and lower zone hazy opacities. There are bilateral hilar and perihilar linear opacities. Bilateral pleural effusions are present. The right pleural effusion is greater than the left. No pneumothorax is observed on the right. No cardiomegaly is present.

\end{quote}
\end{scriptsize}

\textbf{Stage 3: Creating Progression-Specific Reports.}
For each predefined finding of interest, we generate three reports—one each for "improved," "stable," and "worsened" progression. If a specific finding does not exist in the original report, we synthesize the corresponding sentence.
The prompt used to generate these progression-specific sentences is:
\begin{scriptsize}
\begin{quote}
**Role:** You are an expert assistant specialized in processing medical text, specifically Chest X-Ray (CXR) reports.

**Task:** Given a CXR report text and a specific clinical 'finding', perform the following steps:

1.  **Preprocessing:**
    * Review the input CXR report.
    * Remove any sentences that *solely* describe the view position (e.g., "PA and lateral views were obtained.", "AP portable view.", "Single frontal view provided."). Do *not* remove view information if it's integrated into a sentence describing a finding (though this is less common).
    * Retain all sentences describing clinical observations, findings, comparisons, and impressions. Avoid removing other general "unnecessary" words; focus primarily on removing dedicated view position sentences.

2.  **Modification based on Finding:**
    * Identify if the preprocessed report text contains mentions of the provided 'finding'.
    * **If the finding is mentioned:**
        * Locate the primary sentence(s) describing the status or appearance of the 'finding'.
        * Create three versions of the preprocessed report:
            * **Improved:** Modify the relevant sentence(s) minimally to indicate the finding has 'improved', 'decreased', 'resolved', or similar positive change.
            * **Stable:** Modify the relevant sentence(s) minimally to indicate the finding is 'stable', 'unchanged', or 'similar'. If the original text already implies stability, ensure this version reflects that clearly.
            * **Worsened:** Modify the relevant sentence(s) minimally to indicate the finding has 'worsened', 'increased', become 'more severe', or similar negative change.
        * Make *only the minimal changes* necessary to the specific part about the finding's status. Keep the rest of the report text identical to the preprocessed version.
    * **If the finding is NOT mentioned:**
        * Create three versions of the report by appending a new, concise sentence to the end of the preprocessed report text:
            * **Improved:** Append a sentence like: "The [finding] shows improvement." or "[Finding] is improved."
            * **Stable:** Append a sentence like: "The [finding] appears stable." or "[Finding] is stable."
            * **Worsened:** Append a sentence like: "The [finding] has worsened." or "There is worsening of the [finding]." or "[Finding] has increased."
        * Use the original preprocessed report text for the beginning of each version.

3.  **Output:**
    * Format the final output as a single JSON object string.
    * The JSON object must have exactly three keys: `"improved"`, `"stable"`, and `"worsened"`.
    * The value for each key should be the full text of the corresponding modified report generated in Step 2. Ensure the output is valid JSON.

**Input Format Reminder:** The user will provide input in the following format:
`finding`: [The specific clinical finding]
`report`: [The full text of the CXR report]

**Example 1 (Illustrative - do not repeat in output):**
`finding`: pleural effusion
`report`: "When compared to the prior study, the left-sided pleural effusion appears stable. Left consolidation appears relatively stable. No pneumothoraces are seen. The rest of the support lines and tubes are unchanged in position. PA and lateral views were obtained."

**Expected Output Example 1 (Illustrative - do not repeat in output):**
```json
\{
  "improved": "When compared to the prior study, the left-sided pleural effusion is improved. Left consolidation appears relatively stable. No pneumothoraces are seen. The rest of the support lines and tubes are unchanged in position.",
  "stable": "When compared to the prior study, the left-sided pleural effusion appears stable. Left consolidation appears relatively stable. No pneumothoraces are seen. The rest of the support lines and tubes are unchanged in position.",
  "worsened": "When compared to the prior study, the left-sided pleural effusion is worsened. Left consolidation appears relatively stable. No pneumothoraces are seen. The rest of the support lines and tubes are unchanged in position."
\}

**Example 2 (Illustrative - do not repeat in output):**
`finding`: pneumothorax
`report`: "Interval improved aeration is noted at both lung bases. Residual patchy and linear left lower lobe atelectasis remains. A small left pleural effusion is present. Single frontal view."

**Expected Output Example 2 (Illustrative - do not repeat in output):**
```json
\{
  "improved": "Interval improved aeration is noted at both lung bases. Residual patchy and linear left lower lobe atelectasis remains. A small left pleural effusion is present. The Pneumothorax is improved.",
  "stable": "Interval improved aeration is noted at both lung bases. Residual patchy and linear left lower lobe atelectasis remains. A small left pleural effusion is present. The Pneumothorax is stable.",
  "worsened": "Interval improved aeration is noted at both lung bases. Residual patchy and linear left lower lobe atelectasis remains. A small left pleural effusion is present. The Pneumothorax has worsened."
\}

Now process the following input:
\end{quote}
\end{scriptsize}

\section{Binary Interval-Change Dataset}
\label{subsec:interval_binary}

We construct binary interval-change labels (\texttt{change} vs.\ \texttt{no change}) from radiology report impressions across four datasets. Cases are labeled as \texttt{no change} if the impression contains the phrase \textit{``no interval change.''} Cases are labeled as \texttt{change} if the impression contains any of the following progression-related keywords:

\begin{itemize}
    \item \textbf{Worsening:} aggravated, exacerbated, increase, worsen, progression, enlarged
    \item \textbf{Improving:} improve, decrease, diminished, reduce, regress, resolve, disappear
    \item \textbf{New/Developing:} new, newly, developed, developing, recur, recurrence
    \item \textbf{Interval changes:} interval decrease, interval increase, interval improvement, interval worsening
\end{itemize}

\noindent Reports matching neither criterion are excluded. Label counts per dataset are summarized in \cref{tab:binary_labels}.

\begin{table}[h]
\centering
\caption{Binary interval-change label distribution. Label~0 denotes \texttt{no change} and Label~1 denotes \texttt{change}.}
\label{tab:binary_labels}
\begin{tabular}{lrrr}
\toprule
Dataset & Label 0 & Label 1 & Total \\
\midrule
CheXpert    & 49{,}591 & 53{,}505 & 103{,}096 \\
MIMIC       & 22{,}527 & 45{,}370 &  67{,}897 \\
RexGradient &  8{,}902 & 10{,}921 &  19{,}823 \\
Private        &  5{,}000 &  5{,}000 &  10{,}000 \\
\bottomrule
\end{tabular}
\end{table}
\noindent CheXpert and MIMIC exhibit moderate class imbalance toward \texttt{change}, reflecting the higher prevalence of progression-related language in follow-up reports. The private hospital cohort (SNU2) is balanced by randomly sampling 5{,}000 cases per class from the full dataset.

\section{Experiment}
\label{subsec:experiment}

\subsection{Zero-Shot Prompt Design}
For each finding and progression class, we design 12–17 distinct prompts to capture the diverse phrasing typically found in radiology reports. Using multiple prompts per class helps reduce score variance, as relying on a single template can lead to unstable results. During zero-shot classification, we compute the cosine similarity between the image representation and each prompt corresponding to a specific progression label, and average these scores to obtain the final prediction.
\begin{lstlisting}
all_prompt={
    'pneumothorax':
        {'improving':["Improved right pneumothorax.",
            "Decreased size of pneumothorax compared to prior.",
            "Interval improvement in pneumothorax.",
            "Partial resolution of left pneumothorax.",
            "Pneumothorax has decreased in size.",
            "Improvement in previously noted pneumothorax.",
            "Marked reduction in size of pneumothorax.",
            "Smaller right apical pneumothorax noted today.",
            "Improved pneumothorax, no acute findings.",
            "Reduction in pneumothorax volume.",
            "Improved appearance of left apical pneumothorax.",
            "Pneumothorax is resolving.",
            "Pneumothorax shows interval decrease."],
        'stable' : [
            "Stable small right pneumothorax.",
            "No increase in size of pneumothorax.",
            "Pneumothorax appears unchanged from prior.",
            "Small left pneumothorax, no acute findings.",
            "No evidence of expanding pneumothorax.",
            "Pneumothorax is stable with no tension physiology.",
            "Minimal pneumothorax, no intervention needed.",
            "Persistent small pneumothorax without progression.",
            "Pneumothorax noted, patient remains stable clinically.",
            "No signs of worsening pneumothorax.",
            "Pneumothorax is stable in appearance and size.",
            "No interval change in pneumothorax.",
            "Left apical pneumothorax stable compared to prior.",
            "Pneumothorax remains small and non-tension."],
        'worsening' : [
            "Worsening right pneumothorax.",
            "Increased size of pneumothorax compared to prior.",
            "Interval increase in pneumothorax.",
            "Progression of left pneumothorax.",
            "Pneumothorax has enlarged.",
            "Worsening left apical pneumothorax.",
            "Marked increase in pneumothorax size.",
            "Pneumothorax increasing, consider intervention.",
            "Expansion of previously noted pneumothorax.",
            "Pneumothorax now involves greater lung volume.",
            "New increase in size of right pneumothorax.",
            "Pneumothorax shows interval worsening.",
            "Pneumothorax progressing compared to previous imaging.",
            "Enlarging pneumothorax noted on follow-up."]
        },
    'pleural_effusion':{
        "improving": [
            "Improved right pleural effusion.",
            "Decreased size of pleural effusion compared to prior.",
            "Interval improvement in pleural effusion.",
            "Partial resolution of left pleural effusion.",
            "Reduction in pleural effusion volume.",
            "Pleural effusion has decreased in size.",
            "Marked reduction in right pleural effusion.",
            "Pleural effusion is resolving.",
            "Improved appearance of left pleural effusion.",
            "Improvement in previously noted pleural effusion.",
            "Less fluid seen in pleural space than before.",
            "Pleural effusion shows interval decrease.",
            "Decreased right basilar pleural effusion.",
        ],
        "stable": [
            "Stable small right pleural effusion.",
            "No increase in size of pleural effusion.",
            "Pleural effusion appears unchanged from prior.",
            "Small left pleural effusion, no acute findings.",
            "No evidence of expanding pleural effusion.",
            "Pleural effusion is stable in appearance and size.",
            "Minimal pleural effusion, no intervention needed.",
            "Persistent small pleural effusion without progression.",
            "Pleural effusion noted, patient remains stable clinically.",
            "No signs of worsening pleural effusion.",
            "No interval change in pleural effusion.",
            "Left basilar pleural effusion stable compared to yesterday.",
            "Pleural effusion remains small and unchanged.",
            "Stable bilateral pleural effusions."
        ],
        "worsening": [
            "Worsening right pleural effusion.",
            "Increased size of pleural effusion compared to prior.",
            "Interval increase in pleural effusion.",
            "Progression of left pleural effusion.",
            "Pleural effusion has enlarged.",
            "Worsening left basilar pleural effusion.",
            "Marked increase in pleural effusion size.",
            "Expansion of previously noted pleural effusion.",
            "Pleural effusion increasing, consider intervention.",
            "New increase in size of pleural effusion.",
            "Pleural effusion now causes greater lung compression.",
            "Pleural effusion shows interval worsening.",
            "Pleural fluid accumulation appears progressive.",
            "Enlarging pleural effusion noted on follow-up."
        ]
        },
    'consolidation':{
        "improving": [
            "Improved right lower lobe consolidation.",
            "Decreased area of consolidation.",
            "Interval improvement in consolidation.",
            "Consolidation has partially resolved.",
            "Marked reduction in pulmonary consolidation.",
            "Clearing of previously noted consolidation.",
            "Consolidation is less extensive than prior.",
            "Improved left basilar consolidation.",
            "Reduction in airspace consolidation.",
            "Consolidation resolving on follow-up imaging.",
            "Less dense consolidation compared to prior.",
            "Airspace opacity improving.",
            "Consolidation has diminished since prior study.",
            "Patchy consolidation appears improved.",
            "Fading consolidation with treatment."
        ],
        "stable": [
            "Stable consolidation in right lower lobe.",
            "No significant change in consolidation.",
            "Persistent left basilar consolidation.",
            "Consolidation appears unchanged from prior.",
            "Airspace opacity remains stable.",
            "No interval change in consolidation.",
            "Chronic consolidation with no acute findings.",
            "Stable patchy consolidation.",
            "Consolidation noted without progression.",
            "No new consolidation identified.",
            "Findings consistent with stable consolidation.",
            "Consolidation remains unchanged.",
            "Stable appearance of parenchymal consolidation.",
            "Consolidation is chronic and stable.",
            "No worsening of consolidation."
        ],
        "worsening": [
            "Worsening right upper lobe consolidation.",
            "Increased area of consolidation.",
            "Consolidation more extensive than prior.",
            "Interval progression of consolidation.",
            "New or expanding consolidation noted.",
            "Consolidation has worsened.",
            "Marked increase in pulmonary consolidation.",
            "Confluent consolidation involving multiple lobes.",
            "Airspace consolidation increasing.",
            "Progressive dense consolidation.",
            "Patchy consolidation more pronounced.",
            "Worsening consolidation despite treatment.",
            "New left basilar consolidation with progression.",
            "Expanding area of alveolar consolidation.",
            "Increased opacification consistent with worsening consolidation."
        ]
        },
    'edema': {
        "improving": [
            "Improved pulmonary edema.",
            "Decreased pulmonary vascular congestion.",
            "Edema appears less prominent than prior.",
            "Interval improvement in pulmonary edema.",
            "Partial resolution of interstitial edema.",
            "Reduction in alveolar edema.",
            "Pulmonary edema has decreased in extent.",
            "Marked reduction in pulmonary edema.",
            "Clearing of previously seen pulmonary edema.",
            "Improved vascular congestion.",
            "Improved interstitial markings.",
            "Pulmonary edema resolving with treatment.",
            "Decreased perihilar opacities.",
            "Edema improving compared to previous study.",
            "Less pulmonary edema seen on current film."
        ],
        "stable": [
            "Stable pulmonary edema.",
            "No significant change in pulmonary edema.",
            "Pulmonary edema appears unchanged from prior.",
            "Edema remains stable in extent.",
            "Persistent mild pulmonary edema.",
            "Pulmonary vascular congestion unchanged.",
            "Interstitial markings stable.",
            "No interval change in pulmonary edema.",
            "Pulmonary edema noted without progression.",
            "No worsening of pulmonary edema.",
            "Edema appears chronic and stable.",
            "Stable vascular congestion.",
            "No new signs of fluid overload.",
            "Pulmonary edema similar to previous exam.",
            "Mild pulmonary edema, no acute change."
        ],
        "worsening": [
            "Worsening pulmonary edema.",
            "Increased pulmonary vascular congestion.",
            "Edema appears more prominent than prior.",
            "Interval increase in pulmonary edema.",
            "Progressive alveolar edema.",
            "Pulmonary edema has worsened.",
            "Marked increase in pulmonary edema.",
            "Expansion of interstitial edema.",
            "New or worsening bilateral pulmonary edema.",
            "Pulmonary edema now more confluent.",
            "Increasing perihilar opacities.",
            "Worsening interstitial markings.",
            "Pulmonary congestion progressing.",
            "Increased fluid overload signs on imaging.",
            "Diffuse worsening of pulmonary edema pattern."
        ]},
    'pneumonia':{
        "improving": [
            "Improved right lower lobe pneumonia.",
            "Decreased consolidation in left lung.",
            "Pneumonia shows interval improvement.",
            "Clearing of previously seen infiltrates.",
            "Reduction in airspace opacity.",
            "Partial resolution of pneumonia.",
            "Consolidation is less extensive than prior.",
            "Improved left basilar pneumonia.",
            "Decreased right middle lobe opacities.",
            "Pneumonia improving with antibiotic therapy.",
            "Pulmonary infiltrates have diminished.",
            "Improved patchy opacities.",
            "Interval decrease in parenchymal opacities.",
            "Airspace disease appears less prominent.",
            "Pneumonia resolving compared to previous imaging."
        ],
        "stable": [
            "Stable right lower lobe pneumonia.",
            "Pneumonia appears unchanged from prior.",
            "Persistent left basilar consolidation.",
            "No interval change in airspace disease.",
            "Patchy infiltrates stable in appearance.",
            "Consolidation remains without significant change.",
            "No progression of pneumonia.",
            "Airspace opacity unchanged.",
            "Stable pneumonia on follow-up imaging.",
            "No new consolidation identified.",
            "Chronic-appearing infiltrates, no acute change.",
            "Stable bilateral patchy opacities.",
            "No worsening of pneumonia noted.",
            "Findings consistent with prior pneumonia, stable.",
            "Stable airspace disease, no interval change."
        ],
        "worsening": [
            "Worsening right upper lobe pneumonia.",
            "Increased consolidation in left lower lobe.",
            "Pneumonia appears more extensive than prior.",
            "Interval progression of pneumonia.",
            "New or worsening bilateral infiltrates.",
            "Airspace opacities have increased.",
            "Expansion of previously seen pneumonia.",
            "Pneumonia worsening despite treatment.",
            "More confluent consolidation noted today.",
            "Increasing parenchymal opacities.",
            "Marked progression of airspace disease.",
            "Increased patchy opacities compared to prior.",
            "Pulmonary infiltrates have progressed.",
            "Worsening left lower lobe pneumonia.",
            "New consolidation suggestive of worsening pneumonia."
        ]
        }
}
\end{lstlisting}

\subsection{Ablation for Hyperparameters}
\label{sec:abl_hyperparameters}

We ablate the pretraining weight $W$ for the Change-aware Sigmoid loss and the fine-tuning weight $\lambda$ for the Temporal Consistency Loss (TCL). 
Table~\ref{tab:abl_w} reports average macro-accuracy on MS-CXR-T\textsubscript{retrieval} when varying $W$. 
Although $W=0.5$ achieves the highest Standard accuracy, $W=1$ provides the best overall trade-off, yielding the strongest Reversed, Combined, and Consistency scores. 
Setting $W=2$ places too much emphasis on the inverted pairs and degrades performance across all protocols.

Table~\ref{tab:abl_lambda} summarizes the effect of the fine-tuning weight $\lambda$ on supervised MS-CXR-T classification. 
Introducing BiCE alone ($\lambda=0$) already results in a substantial improvement over the baseline across all evaluation protocols, demonstrating that enforcing label inversion provides a strong directional signal. 
Adding TCL with a small weight ($\lambda=1$) produces performance similar to $\lambda=0$, reflecting the scale difference between the cross-entropy and TCL objectives. 
Increasing the TCL weight to $\lambda=50$ yields the best balanced performance, with the high Reversed, Combined, and Consistency scores and only a minor reduction relative to the BiCE-only setting in the Standard metric. 
At $\lambda=100$, directional metrics remain strong, but Standard and Combined accuracies begin to drop, indicating that overly large TCL weights may over-regularize the model. 
Based on these trends, we adopt $\lambda=50$ for all main experiments.

\begin{table}[h]
\centering
\caption{Effect of pretraining weight $W$ on \textbf{MS-CXR-T\textsubscript{retrieval}}. 
We report average macro-accuracy (\%) across all findings for each evaluation protocol.}
\label{tab:abl_w}
\renewcommand{\arraystretch}{1.1}
\begin{tabular}{ccccc}
\toprule
\multirow{2}{*}{$W$} & \multicolumn{4}{c}{Average Accuracy (\%)} \\
\cmidrule(lr){2-5}
 & Standard & Reversed & Combined & Consistency \\
\midrule
0   & 50.2 & 50.9 & 55.4 & 32.7 \\
0.5 & \textbf{54.9} & 53.2 & 59.1 & 36.6 \\
1   & 54.1 & \textbf{54.0} & \textbf{59.2} & \textbf{37.9} \\
2   & 49.7 & 51.2 & 54.3 & 33.1 \\
\bottomrule
\end{tabular}
\end{table}

\begin{table}[h]
\centering
\caption{Effect of fine-tuning weight $\lambda$ on \textbf{MS-CXR-T} supervised classification. 
We report average macro-accuracy (\%) across all findings for each evaluation protocol.}
\label{tab:abl_lambda}
\renewcommand{\arraystretch}{1.1}
\begin{tabular}{ccccc}
\toprule
\multirow{2}{*}{$\lambda$} & \multicolumn{4}{c}{Average Accuracy (\%)} \\
\cmidrule(lr){2-5}
 & Standard & Reversed & Combined & Consistency \\
\midrule
Base   & 61.1 & 53.3 & 59.6 & 39.5 \\
0   & \textbf{65.2} & 62.0 & 63.4 & 53.1 \\
1   & 65.0 & 61.2 & 63.1 & 53.3 \\
50  & 64.1 & 63.7 & \textbf{63.6} & \textbf{57.3} \\
100 & 62.7 & \textbf{63.8} & 61.4 & 55.6 \\
\bottomrule
\end{tabular}
\end{table}

Also, in practice, we do not apply the inversion-aware objectives from the very beginning of training. 
If BiCE or TCL is introduced too early, the model can quickly converge to a degenerate solution that predicts \texttt{stable} for most pairs, which locally minimizes the bidirectional losses but is not clinically meaningful. 
To avoid this shortcut, we first warm up the model with standard objectives and then enable the Change-aware Sigmoid loss after 10 pretraining epochs, and TCL after 20 epochs of fine-tuning. 
This staging allows the backbone to learn non-trivial temporal structure before inversion-aware regularization is applied.

\section{Additional Information}
\label{subsec:etc}

\subsection{Clinical Utility of Reversed and Combined Evaluations}
\label{subsec:clinical_utility}
Radiologists typically assess temporal progression only in the forward (standard) direction. The \textit{Reversed} and \textit{Combined} evaluations are therefore introduced as analytical tools to rigorously validate the reliability of forward predictions. 

The \textit{Reversed} evaluation tests whether a model truly understands temporal progression. For instance, a model reaching 70\% accuracy in the forward direction but only 40\% in the reversed direction likely exploits spurious cues rather than genuine temporal reasoning. Such discrepancies may undermine the trustworthiness of forward predictions. 

The \textit{Combined} evaluation complements this by enforcing consistency across both directions. By aggregating results from forward and reversed pairs, it exposes and helps mitigate directional biases. Together, these settings provide crucial analytical validation, ensuring that predictions in standard clinical scenarios are both reliable and interpretable.

\paragraph{Implications for Temporal Inversion.}
While reversible scenarios predominate across these findings, temporal inversion is not intended to model exact biological recovery. Instead, it provides a controlled perturbation that tests whether models capture the \emph{direction} of temporal change. Radiologic follow-up primarily evaluates changes in lesion size, burden, or conspicuity—features that exhibit approximate reversibility and therefore lend themselves to inversion-based analysis. Our experiments Our experiments indicate that including inverted pairs improves directional sensitivity without compromising forward-order predictions, supporting temporal inversion as a practical stress test for assessing order-aware interval-change modeling.

\subsection{Use of Large Language Models}
We disclose the use of large language models (LLMs) in this work. 
LLMs were used in three ways: 
(i) to refine the clarity and presentation of writing; 
(ii) to assist in generating change/no-change labels from radiology reports (see \cref{subsec:changelabel}); 
and (iii) to construct the MS-CXR-T\textsubscript{retrieval} benchmark by modifying radiology reports under controlled prompts (see \cref{subsec:mscxrtretrieval}). 
All LLM usage was limited to these supporting roles and did not alter the core experimental results or conclusions.

\end{document}